\documentclass{article}
\usepackage{ijcai23}

\usepackage{times}
\usepackage{soul}
\usepackage{url}
\usepackage[hidelinks]{hyperref}
\usepackage[utf8]{inputenc}
\usepackage[small]{caption}
\usepackage{graphicx}
\usepackage{amsmath}
\usepackage{amsthm}
\usepackage{booktabs}
\usepackage{algorithm}
\usepackage{algorithmic}
\usepackage[switch]{lineno}
\usepackage{multirow}
\usepackage{makecell}
\usepackage{pifont}
\usepackage{enumitem}

\title{A Multi-Modal Neural Geometric Solver with Textual Clauses \\ Parsed from Diagram}

\author{
Ming-Liang Zhang$^{1,2}$\and
Fei Yin$^{1,2}$\and
Cheng-Lin Liu$^{1,2}$\\
\affiliations
$^1$State Key Laboratory of Multimodal Artificial Intelligence Systems, Institute of Automation of Chinese Academy of Sciences\\
$^2$School of Artificial Intelligence, University of Chinese Academy of Sciences\\
\emails
zhangmingliang2018@ia.ac.cn,
fyin@nlpr.ia.ac.cn,
liucl@nlpr.ia.ac.cn
}

\begin{document}

\maketitle

\begin{abstract}
Geometry problem solving (GPS) is a high-level mathematical reasoning requiring the capacities of multi-modal fusion and geometric knowledge application. Recently, neural solvers have shown great potential in GPS but still be short in diagram presentation and modal fusion. In this work, we convert diagrams into basic textual clauses to describe diagram features effectively, and propose a new neural solver called PGPSNet to fuse multi-modal information efficiently. Combining structural and semantic pre-training, data augmentation and self-limited decoding, PGPSNet is endowed with rich knowledge of geometry theorems and geometric representation, and therefore promotes geometric understanding and reasoning. In addition, to facilitate the research of GPS, we build a new large-scale and fine-annotated GPS dataset named PGPS9K, labeled with both fine-grained diagram annotation and interpretable solution program. Experiments on PGPS9K and an existing dataset Geometry3K validate the superiority of our method over the state-of-the-art neural solvers. Our code, dataset and appendix material are available at \url{https://github.com/mingliangzhang2018/PGPS}.
\end{abstract}

\section{Introduction}
Automatic geometry problem solving (GPS) is a long-standing and challenging AI task, and has attracted much attention in the CV and NLP community recently \cite{Seo2015,Sachan2017,Lu2021,Chen2021,Zhang2021}. A geometry problem is formed with a textual problem and a geometry diagram, where the textual problem describes the geometry problem condition and sets the reasoning objective in natural language, and the geometry diagram carries rich structural and semantic information beyond the textual problem to aid problem solving. GPS requires mathematical and multi-modal reasoning capabilities combining the textual problem and geometry diagram. The multi-modal integration and geometric knowledge application are the keys of GPS. 


\begin{figure}[t]
    \begin{center}
    \includegraphics[width=0.9 \columnwidth]{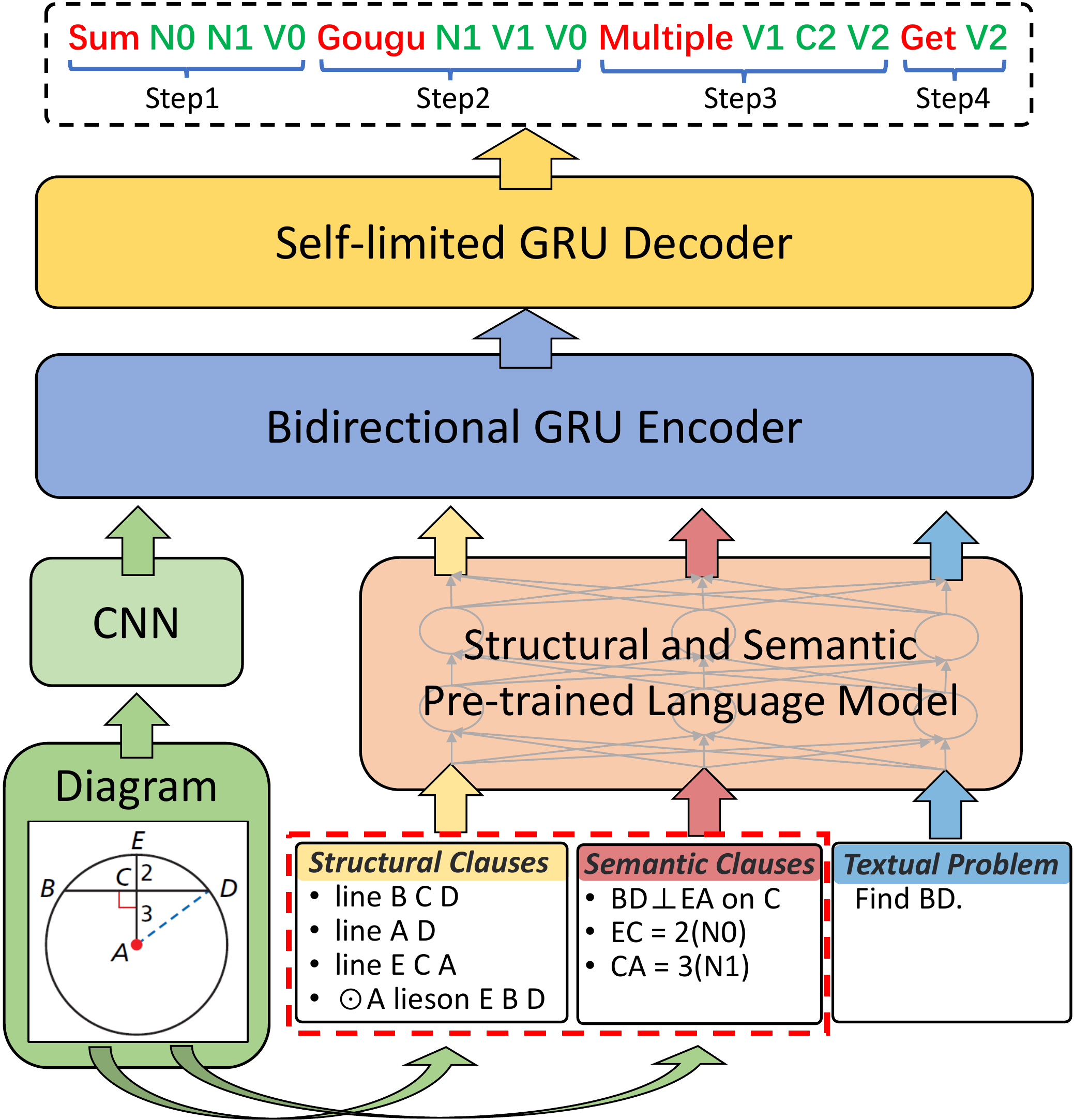} 
    \end{center}
    \vspace{-0.1cm}
    \caption{Overview of PGPSNet solver. PGPSNet is a multi-modal learning framework whose modal inputs contain not only the diagram and textual problem, but also the textual clauses parsed from diagram. It generates the theorem-based interpretable solution program to solve geometry problem.}
    \label{PGPSNet}
    \vspace{-0.2cm}
\end{figure}



Existing works of GPS can be classified into two categories: symbolic geometric solvers \cite{Seo2015,Sachan2017,Lu2021} and neural geometric solvers \cite{Chen2021,cao2022,Chen2022}. The \textit{symbolic solvers} parse the diagram and textual problem into a unified formal language. Based on the geometric theorem knowledge, they usually perform symbolic reasoning by path search and condition matching to produce new conditional states until they find the search target. While the symbolic solvers have better interpretability compared with neural solvers, they are carefully designed with complex rules and hard to extend. Besides, some symbolic solvers may solve problems slowly with many redundant steps, and the search process also does not match humans' solutions. The \textit{neural solvers} proposed recently embed the diagram and textual problem jointly with the hybrid encoder and self-supervised auxiliary tasks and generate the solution program in sequential form. Although neural solvers have achieved impressive results with simple pipelines, recent works \cite{Lu2021,Lu2022ASO} show that there remains a large performance gap compared with symbolic solvers. One of major reasons is that neural solvers, adopting similar frameworks of general multi-modal reasoning tasks applied for natural images, cannot exploit the diagram efficiently. Because primitives in geometry diagram are slender and overlapped, and have complex spatial relationship, the feature map based \cite{Peter2018}, region proposal based \cite{Zhou2019} or image patch based \cite{Kim2021} frameworks cannot extract fine-grained features and even damage structural and semantic information. 

Considering the under-representation of diagram and difficulty of cross-modal fusion in neural solver, we represent the geometry diagram by textual clauses including structural clauses and semantic clauses, as demonstrated in bottom of Figure \ref{PGPSNet}. Compared with visual image, clauses have highly syntactic structures with less redundant information naturally. Textual clauses are better suited to describe fine-grained and multi-level information in geometry diagram. Different from the formal language consisting of complex multi-order logic forms in symbolic solver, our clauses only describe basic relationships, where structural clauses depict the connection relations among geometric primitives and the semantic clauses describe the semantic relations between non-geometric primitives and geometric primitives. We do not pursue higher level relations constructed by geometric rules since they do not conform to the goal of neural solver. The model is hoped to have the ability of matching geometric patterns and constructing high-level relations from basic relations. 

To promote the research on GPS, we also build a new large-scale and fine-annotated GPS dataset named PGPS9K. PGPS9K contains 9,022 geometry problems paired with geometry diagrams. In contrast to existing datasets, PGPS9K is labeled with both diagram annotation and solution program. The diagram annotation employs the same primitive level annotation method as \cite{Zhang2021,Hao2022PGDP5KAD}, which could be translated to textual clauses simply and uniquely. Given the solution complexity of GPS, we design a new annotation form based on geometric theorems for solution program. The solution program provides the problem solving procedure wherein each step is an application of a theorem (axiom), as shown in top of Figure \ref{PGPSNet}, rather than the specific solution steps involving fundamental arithmetic operations \cite{wang-etal-2017-deep,amini-etal-2019-mathqa}. Our solution program carries rich geometric knowledge and has better interpretability, and as well reduces the burden of model learning.

Taking advantage of multi-modal information, we thus propose a new diagram-text fusion neural solver named PGPSNet as shown in Figure \ref{PGPSNet}. In addition to the geometry diagram and textual problem, PGPSNet combines structural clauses and semantic clauses and generates the solution program to solve problem. To fuse different parts of text modality, we propose a structural and semantic pre-training strategy based on Masked Language Modeling (MLM) task \cite{bert2019}, for improving the model's understanding of structural and semantic content by explicit modeling. To overcome the limitation of data size and pre-trained corpus, we design five data augmentation strategies based on diversity and equivalence of geometric representation. Besides, we construct a self-limited GRU decoder to shrink the representation space and search space of operands and speed up training and inference. Experiments on an existing dataset Geometry3K \cite{Lu2021} and PGPS9K dataset demonstrate that our PGPSNet boosts the performance of GPS prominently, largely exceeds the performance of existing neural solvers, and also achieves comparable results as well-designed symbolic solvers.

The contributions of this work are summarized in four folds: (1) We use textual clauses to express the fine-grained structural and semantic information in geometry diagram efficiently. (2) We propose a new neural solver PGPSNet, fusing multi-modal information through structural and semantic pre-training, data augmentation, and self-limited decoding. (3) We construct a large-scale dataset PGPS9K labeled with both fine-grained diagram annotation and interpretable solution program. (4) Our PGPSNet outperforms existing neural solvers significantly. 

\begin{table*}
    \centering
    \footnotesize
    \renewcommand\arraystretch{1.1}
    \begin{tabular}{lccccccc}
        \toprule
        Dataset    & \#QA & Grade & \#Type & Diagram Anno & Rationale     & \#Avg OP & \#Avg PL  \\
        \midrule
        GEOS \cite{Seo2015} & 186  & 6-10 & - & No & - & - & - \\
        GEOS++ \cite{Sachan2017} & 1,406 & 6-10 & - & No & - & - & - \\
        GEOS-OS \cite{sachan-xing-2017-learning} & 2,235 & 6-10 & - & No & Demonstration & - & - \\
        Geometry3K \cite{Lu2021} & 3,002 & 6-12 & 4 & Yes & Logical form  & - & - \\
        GeoQA \cite{Chen2021} & 4,998 & 6-12 & 3 & No & Program & 1.98 & 5.35 \\
        GeoQA+ \cite{cao2022} & 7,528 & 6-12 & 3 & No & Program & 2.61 & - \\
        PGPS9K & 9,022 & 6-12 & 30 & Yes & Program & 2.43 & 7.45  \\
        \bottomrule
    \end{tabular}
    \caption{Comparison with existing GPS datasets. Type, OP and PL represent problem type, operator number and program length, respectively.}
    \label{dataset_compare}
\end{table*}

\begin{figure*}
    \begin{center}
    \includegraphics[width=1.65\columnwidth]{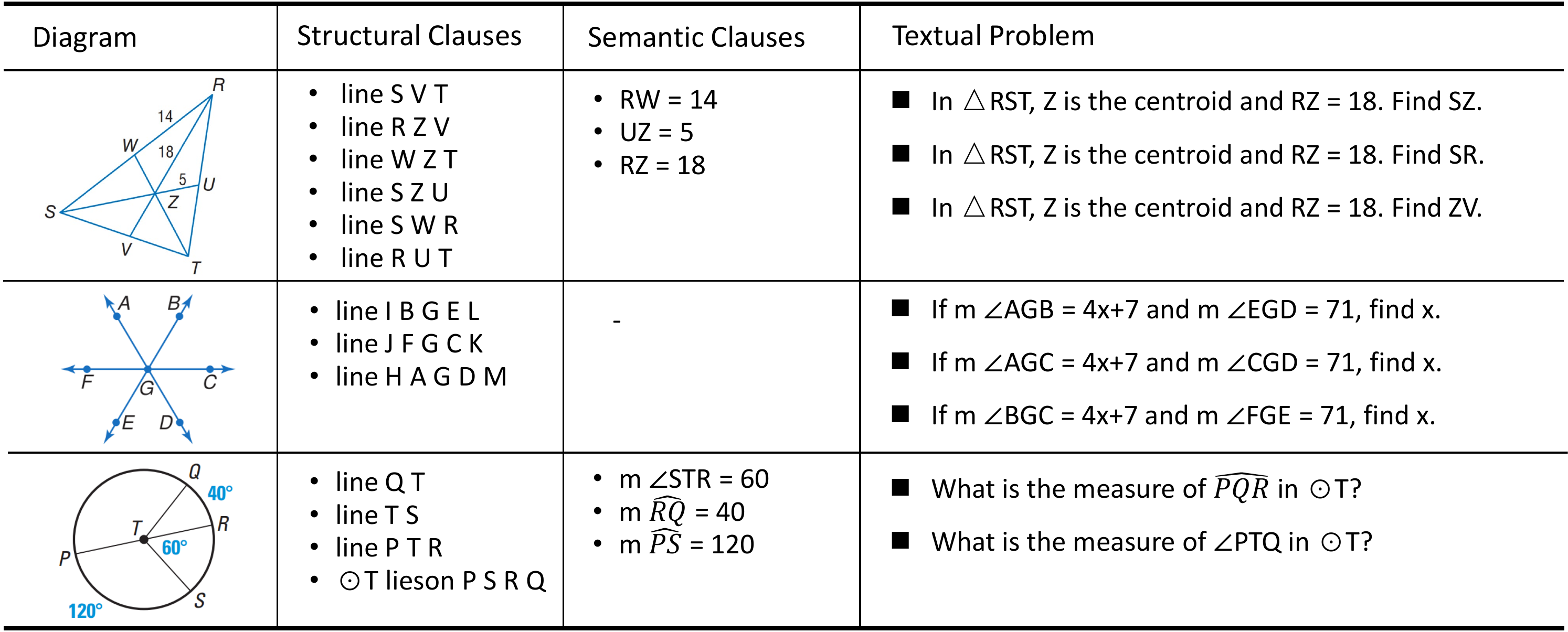} 
    \end{center}
    \vspace{-0.2cm}
    \caption{Example presentation of PGPS9K dataset.}
    \label{problem_example}
    \vspace{-0.3cm}
\end{figure*} 

\section{Related Work}
\subsection{Multi-modal Reasoning}
Multi-modal reasoning uses multi-modal datasets to perform reasoning tasks, e.g., visual question answering \cite{Peter2018,Kim2021}, document visual question answering \cite{Xu2020LayoutLMPO,Tito2021}, table question answering \cite{zhu-etal-2021-tat,lu2022dynamic}, where GPS is also a special multi-modal reasoning task. Because of difference of data modality and reasoning ability, it results in significant semantic gaps between different tasks. The core to multi-modal reasoning lies on how to unite modalities and incorporate domain knowledge. 

\subsection{Geometry Problem Solving}
Generalized GPS contains geometry calculation \cite{Seo2015,tsai-etal-2021-sequence} and geometry proving \cite{Chou1996,Gan2019}, and can be divided into symbolic solvers \cite{Seo2015,Lu2021} and neural solvers \cite{Chen2021,Chen2022} in method. The symbolic solvers that have been studied for years possess their advantages and limitations as introduced in Introduction. With development of neural network, neural solvers have shown their potential in GPS, whereas they still cannot handle modal representation and fusion well. In context of the geometry calculation, our work unifies the geometry diagram and textual problem into the text modality to better construct structural and semantic relationships in geometry.

\subsection{Pre-trained Language Model for Mathematical Reasoning}
Language models \cite{lewis-etal-2020-bart,brown2020language}, pre-trained on large text corpus with self-supervised learning tasks such as MLM \cite{bert2019} or CLM \cite{lewis-etal-2020-bart}, have demonstrated remarkable performance gains on wide range of NLP tasks such as text classification \cite{Shervin2021} and question answering \cite{khashabi-etal-2020-unifiedqa}. Inspired by them, pre-trained language model is also applied to mathematical reasoning task gradually, e.g., math word problem solving \cite{liang-etal-2022-mwp,zhang2022elastic} and GPS \cite{Lu2021,Chen2022}. However, lacking of large-scale mathematical corpus and targeted pre-training tasks, existing language models pre-trained on natural corpus seem to have limited effects on downstream reasoning tasks with only fine-tune tactic. In this work, we pre-train language model with MLM task combining textual clauses and textual problem, equipping model with capacity of structural and semantic understanding in geometry and therefore improving the performance of GPS substantially.

\section{PGPS9K Dataset}
Although several datasets \cite{Seo2015,Sachan2017,Lu2021,Chen2021} for GPS have been proposed, as presented in Table \ref{dataset_compare}, they are either in small scale only for rule-based symbolic solvers, or coarse-grained annotated neglecting rich information in diagram. To facilitate the application of neural solver, we build a new large-scale GPS dataset called PGPS9K \footnote{\url{http://www.nlpr.ia.ac.cn/databases/CASIA-PGPS9K}} labeled both fine-grained diagram annotation and interpretable solution program. To the best of our knowledge, PGPS9K is the largest and the most complete annotation dataset for GPS up to now.
\subsection{Collection and Description}
PGPS9K is composed of 9,022 geometry problems paired with non-duplicate 4,000 geometry diagrams, where 2,891 problems paired with 1,738 diagrams are selected from Geometry3K dataset \cite{Lu2021}, the rest of problems are collected from five popular textbooks across grades 6-12 on mathematics curriculum websites\footnote{\url{https://www.mheducation.com/}}. Our PGPS9K is divided into 30 problem types as exhibited in appendix A.1, covering almost all problem types of plane geometry problem in corresponding grades. As shown in Figure  \ref{problem_example}, PGPS9K dataset has five properties: (1) \textbf{Theorem-based}: Solving problems in PGPS9K need to apply geometric theorem knowledge to carry out algebraic calculation and get numerical results finally; (2) \textbf{Diagram-dependent}: Above 90\% of problems must be solved using the diagrams because necessary conditions such as variable content and geometric structure are displayed via visual form instead of text; (3) \textbf{Abstract}: The diagram is integrated with basic geometric primitives (point, line, circle) and non-geometric primitives (text, symbol). No complex semantic scenarios are involved in textual problem except abstract geometric conditions; (4) \textbf{Fine-grained}: Problems with the same diagram vary in conditions or targets. Slight distinctions in textual problems usually lead to completely different solutions to problems; (5) \textbf{Condition-redundancy}: Lots of conditions in semantic clauses or textual problem are not needed in problem solving at hand. These five properties make PGPS9K focus on the challenges at geometric reasoning and alleviate the bias introduced by the text \cite{Manjunatha2019,patel-etal-2021-nlp}. Moreover, for convenience of experimental comparison, we split PGPS9K in two ways: The first is leaving out the test set of Geometry3K \cite{Lu2021} as test set (589) and other disjoint samples as training set (8,433); The second is dividing samples of each problem type according to ratio of 8:1 (training set 8,022 and test set 1,000). 

\begin{figure}
    \begin{center}
    \includegraphics[width=0.85\columnwidth]{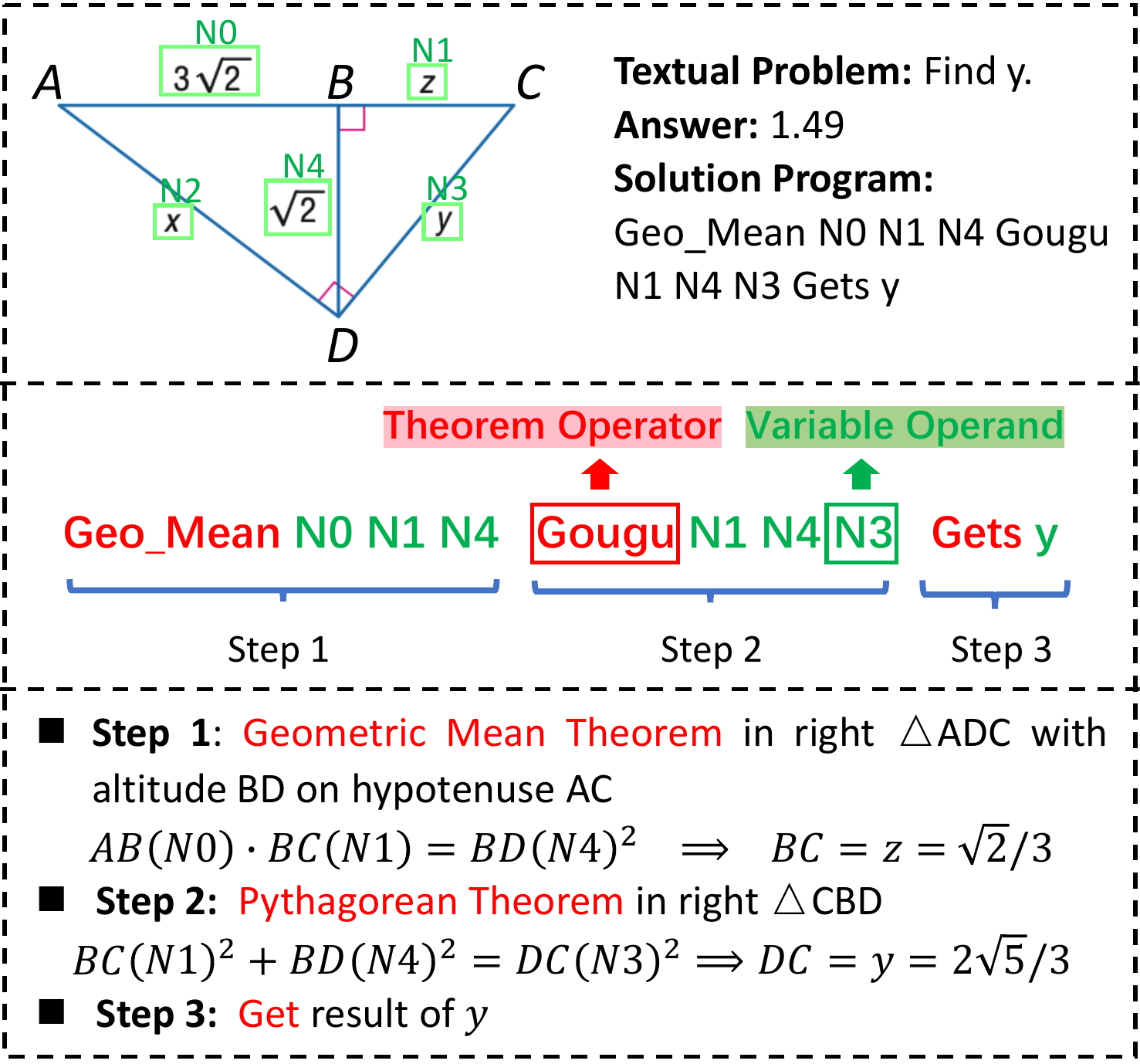} 
    \end{center}
    \vspace{-0.1cm}
    \caption{Annotation of solution program and its interpretability.}
    \label{annotation}
    \vspace{-0.2cm}
\end{figure}

\subsection{Annotation Form}
The annotations of PGPS9K include diagram annotation and solution program, where the diagram annotation is to extract structural and semantic information in diagram and the solution program defines the solution steps of problem. 

Diagram annotation adopts the same primitive level labels as \cite{Zhang2021} which includes primitive contents and primitive relations in tuple form. Then we translate them into two kinds of textual clauses: structural clauses and semantic clauses. The structural clauses are confined to the connection relationship among geometric primitives and described by clauses with points on lines or points on circles, wherein points are arranged in order. The connection relation reveals the most fundamental structural relation displayed in diagram but omitted in textual problem. The semantic clauses depict basic relations between geometric primitives and non-geometric primitives with natural language. These relations are necessary parts for problem solving and complement each other in diagram and textual problem. Noting that the definition and descriptive approach of textual clauses remain open and the overall design principle is to characterize complete features of diagram to help with GPS. More details about textual clauses please refer to appendix A.2. 

Our solution program gives the geometric solution procedure consisting of several deduction steps. It is composed of 34 \textit{operators} $OP$ and 55 \textit{operands} $PN$, where a operator and a few of related operands form one step. Each operator implies one geometric theorem or axiom wherein operands involved are sorted according to the corresponding theorem formula. Operands can be divided into four types: \textit{problem variables} $N$ presented in textual problem or semantic clauses, \textit{process variables} $V$ generated during the process, \textit{arguments} $ARG$ are alphabetic unknowns $[a-z]$, and \textit{constants} $C$. For example, the Pythagorean theorem reveals the relationship of right sides and hypotenuse in right triangle with theorem formula $a^2+b^2=c^2$, so we express it as "Gougu($a$, $b$, $c$)". Compared with other annotation methods proposed \cite{amini-etal-2019-mathqa,tsai-etal-2021-sequence,Chen2021}, our annotation eliminates elementary arithmetic operations such as +, -, *, /, and thus has advantages of structuralization, knowledge-guiding and interpretability (as illustrated in Figure \ref{annotation}), making the solution program more concise and reducing the difficulty of model learning. Besides, we firstly introduce \textit{process variables} $V$ as unknown variables in intra-step and as transfer variables in inter-step, unifying the forward and reverse operations within one theorem. For instance, in the Pythagorean theorem, "Gougu($V$, *, *)" and "Gougu(*, *, $V$)" can be set to solve the right side and hypotenuse, respectively. Paired with annotation form of solution program, we also create a powerful program executor to compute numerical results. It implements symbolic algebraic operations with multiple unknowns according to theorem formulas based on SymPy library of Python. More details of solution program are demonstrated in Appendix A.3. 

\section{PGPSNet Model}

\begin{figure*}
    \begin{center}
    \includegraphics[width=1.90\columnwidth]{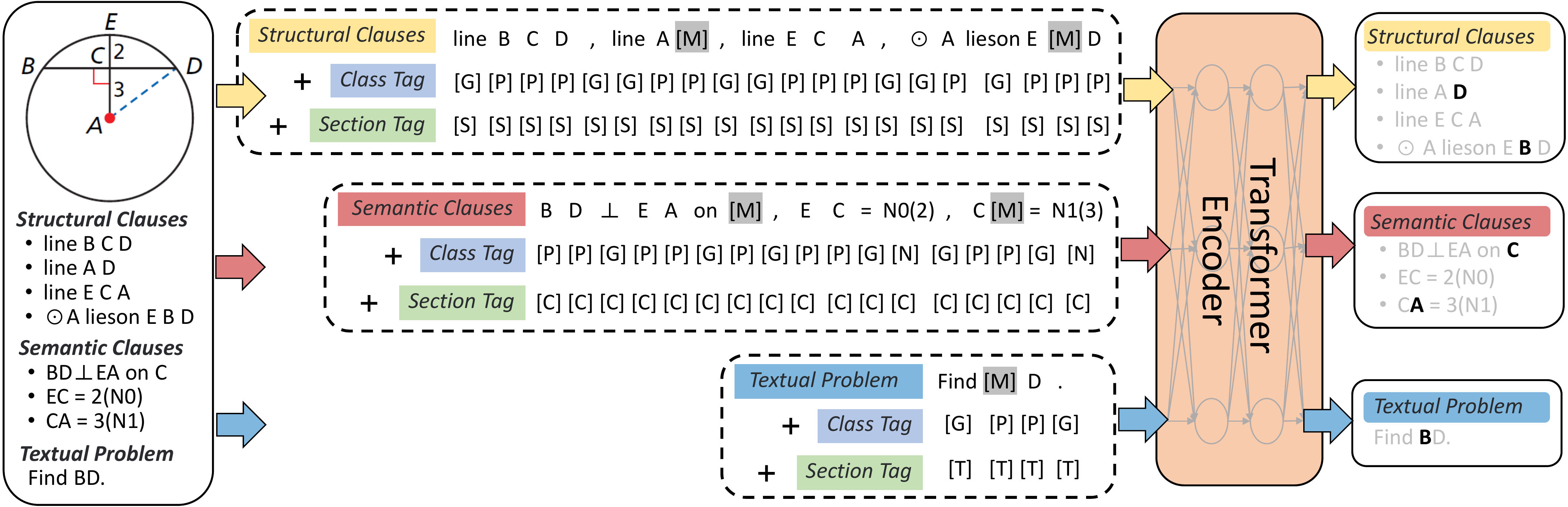} 
    \end{center}
    \caption{Pipeline of structural and semantic pre-training. [M] denotes the mask token. Class tags of [G], [N], [ARG], [P], [ANGID] represent tokens of general, variable, argument, point and angle ID, respectively. Section tags of [S], [C], [T] refer to tokens of structure, condition and target, respectively.}
    \label{Pre-trained}
    \vspace{-0.2cm}
\end{figure*} 

\subsection{Overall Framework}
To fully fuse multi-modality of geometry problem, we propose a new neural solver PGPSNet as depicted in Figure \ref{PGPSNet}. The inputs of PGPSNet include the geometry diagram $D$ and problem text $T$, where the problem text consists of structural clauses $T_{stru}$, semantic clauses $T_{sem}$ and textual problem $T_{prob}$ that $T=\left\{T_{stru}, T_{sem}, T_{prob}\right\}=\left\{t_i\right\}_{i=1}^n$. The diagram image is encoded with convolutional neural network (CNN) and the problem text is passed through the structural and semantic pre-trained language model. Then these two modal tokens are concatenated together, fed into the bidirectional GRU encoder to perform fusing encoding. Next they are decoded by the self-limited GRU decoder to get the solution program $Y=\left\{y_i\right\}_{i=1}^m$ correspondingly. Finally the solution program is calculated by the program executor and obtain the numerical result of geometry problem.

\subsection{Structural and Semantic Pre-training}
While textual clauses describe the fine-grained structural and semantic information parsed from diagram, these clauses are low-level and lack of overall structure and context. Additionally, long, trivial and disordered texts still bring great difficulty to modal fusion and semantic comprehension. Inspired by the pre-training language model, we design a structural and semantic pre-training method based on masked LM task \cite{bert2019} for the text modality learning as shown in Figure \ref{Pre-trained}. 

Firstly, we assign the class tag and section tag for each token in text. The class tag indicates the semantic class of tokens in five classes: general, variable, argument, point and angle ID. The section tag refers to section that tokens belong to, where a problem is divided into structure, condition and target three sections. The input embedding $e_i$ of language model fuses not only positional encoding but also embedding of class tag and section tag as: 
\begin{equation}
    \begin{aligned}
            e_i = TokenEmb(t_i) & + PosEmb(i) + ClassEmb(t_i) \\
            & + SectEmb(t_i), \quad 1 \leq i \leq n. 
    \end{aligned}
\end{equation}
Fine-grained class tags and section tags promote the modeling of geometry problems and alleviate the imbalance of textual token. Then we mask 30\% of text tokens with mask tokens following \cite{cho2021vlt5} but keep tags unchanged. The model is trained to recover the masked text in a unified text generation manner. 

This pre-training method is extremely applicable to structural and semantic modeling of geometry under our modal representation. For instance, it can be reasoned that the mask token in the semantic clause "BD $\perp$ EA on [M]" is "C" according to structural clauses "line B C D" and "line E C A", facilitating model learning the geometric knowledge of line intersection. However, in some cases, model may cannot infer the mask content exactly but geometric knowledge is filled in its token candidates. Taking instance of the textual problem "Find [M]D.", there is high probability that the mask token is "C" or "B" according to the structural clause "line B C D". In summary, pre-training makes model acquire advanced geometric cognition that is an elementary skill for GPS. 

\subsection{Encoder and Decoder}
In our model, the CNN encoder only extracts coarse-grained global visual features of diagram such as geometric style, to determine possible operations quickly and accelerate model convergence. The GRU encoder integrates the diagram encoded as one visual token and textual tokens enhanced by structural and semantic language model, and obtains the mixed encoding context $H^E=\left\{h_i^E\right\}_{i=1}^{n+1}$ as output. 

Because of complexity and flexibility of solution process of geometry problem, solution program cannot convert into binary or general expression tree. Tree decoders widely used in math word problem \cite{ijcai2019p736,tsai-etal-2021-sequence} are not applicable to GPS. We propose a self-limited GRU decoder to generate the sequential solution program in an autoregressive manner. The differences between self-limited decoder and the general attention-based decoder \cite{Bahdanau2014NeuralMT} are two folds: (1) \textit{Self-limited decoder reduces token embedding space.} The input embeddings of problem variables $N$ and augments $ARG$ presented in problem text are copied from encoder output, which also enriches decoder inputs with contextual semantic. Specifically, token embeddings are defined as:
\begin{equation}
    e(y) = \left\{
    \begin{aligned}
        &TokenEmb(y), \quad y \in \{\mathcal{V}_V, \mathcal{V}_C\}, \\
        &h^{E}_{loc(y,T)}, \qquad \qquad  y \in \{\mathcal{V}_N, \mathcal{V}_{ARG}\}, \\
    \end{aligned}
    \right.
\end{equation}
where $\mathcal{V}_V$, $\mathcal{V}_C$, $\mathcal{V}_N$ and $\mathcal{V}_{ARG}$ are target vocabularies of process variables, constants, problem variables and augments, $loc(y,T)$ is the location of token $y$ in problem text $T$. (2) \textit{Self-limited decoder narrows the search space of output tokens.} It limits the output candidates of problem variables $N$ and augments $ARG$ into that appear in the problem text. Concretely, the probability of predicted token $y$ is:
\begin{equation}
           P(y) = Softmax(Score(h^D, c, e(y))),
\end{equation}
where $y\in\{\mathcal{V}_V, \mathcal{V}_C, \mathcal{V}_N \cap T, \mathcal{V}_{ARG} \cap T\}$, $Score$ is the score function, $h^D$ is the hidden output of decoder, $c$ is the context vector generated from $H^E$ using the same attention mechanism as \cite{Bahdanau2014NeuralMT}, $e(y)$ is the token embedding of candidates. In experiments, We find that the self-limited GRU achieves even better performance than complex tree decoders and with much faster training and inference speed.  

\begin{table*}
\centering
\footnotesize
\renewcommand\arraystretch{1.1}
    \begin{tabular}{lcccccc}
        \toprule
        \multirow{2}{*}{Method}  & \multicolumn{3}{c}{Geometry3K} & \multicolumn{3}{c}{PGPS9K} \\
        \cmidrule(lr){2-4} \cmidrule(lr){5-7}
                    & Completion & Choice & Top-3 & Completion & Choice & Top-3  \\
        \midrule
        Human Expert \cite{Lu2021}   & - & 90.9 & - & - & - & - \\
        Baseline (Neural Solver) \cite{Lu2021} & - & 35.9 & -  & - & - & -  \\
        InterGPS (Predict)* \cite{Lu2021} & 44.6  & 56.9  & -  & -   & -  & -  \\
        InterGPS (Diagram GT)* \cite{Lu2021} & 64.2  & 71.7 & - & 59.8 & 68.0 & - \\
        InterGPS (All GT)* \cite{Lu2021} & \textbf{69.0} & 75.9 & - & -  & -  & - \\
        $\text{NGS}^\#$ \cite{Chen2021} & 35.3 & 58.8 & 62.0 & 34.1 & 46.1 & 60.9 \\
        $\text{Geoformer}^\#$ \cite{Chen2022} & 36.8 & 59.3 & 62.5 & 35.6 & 47.3 & 62.3 \\
        PGPSNet  & 65.0 & \textbf{77.9} & \textbf{80.7}  & \textbf{62.7} & \textbf{70.4} & \textbf{79.5}    \\
        \bottomrule
    \end{tabular}
\caption{Numerical answer accuracies (\%) of state-of-the-art GPS solvers. * denotes results re-produced with the authors' code. $\#$ denotes methods re-implemented by us.}
\vspace{-0.3cm}
\label{performance_compare}

\end{table*}

\section{Data Augmentation}
Despite that PGPS9K is the largest dataset so far and of high-quality, it still cannot satisfy the model learning of PGPSNet well, especially for the structural and semantic pre-training task. Therefore, we adopt five data augmentation strategies based on diversity and equivalence of geometric representation, taking the problem in Figure \ref{Pre-trained} as an illustrative example:
\begin{itemize}[leftmargin=*]
	\item[$\bullet$] \textit{Token Replacement}: The replaceable tokens include points, angle IDs and augments three types. Once a token is changed, all same tokens in textual clauses and textual problem should be replaced uniformly. Point B is replaced as point V, then get new texts: "line V C D", "$\odot$A lieson E V D", "VD $\perp$ EA on C", "Find VD". 
    \item[$\bullet$] \textit{Connection Rotation}: The connection relationship in structural clauses could be re-represented by changing the location order of points. "line B C D" is equivalent to "line D C B" with the opposite order, "$\odot$A lieson E B D" is the same as "$\odot$A lieson E D B" in clockwise order. 
    \item[$\bullet$] \textit{Representation Transposition}: There are several equivalent representations of geometric primitives of line, angle and arc, e.g., "EA = AE", "$\angle$STR = $\angle$RTS", "$\widehat{\rm{EF}}$=$\widehat{\rm{FE}}$". We randomly transpose the geometric primitives representation in semantic clauses.  
    \item[$\bullet$] \textit{Clauses Shuffle}: We shuffle semantic clauses to produce new ID of problem variable while modify the corresponding solution program. When semantic clauses are adjusted as "CA = 3(N0), BD $\perp$ EA on C, EC = 2(N1)", the solution program is changed as "$\cdots$ Gougu N0 V1 V0 $\cdots$".
    \item[$\bullet$] \textit{Diagram Flip}: Since the textual content is already parsed in semantic clauses, the visual text in diagram could be ignored. So the flipped or rotated diagram is identical to the original diagram for problem.
\end{itemize}
These five augmentation strategies are independent and can be incorporated with each other. Abundant samples generated from data augmentation equip model with basic geometric knowledge, thus promoting high-level reasoning. 

\section{Experiment}
\subsection{Setup}
\textbf{Implementation details.} \
Our model is implemented using Pytorch on one GTX-RTX GPU. The CNN model adopts the ResNet10 \cite{he2016deep}, feeding with diagram images resized as $128\times128$. The language model select a transformer encoder \cite{Vaswani2017AttentionIA}, having 6 layers, 8 attention heads, and a hidden embedding size of 1024. The GRU encoder is a two-layer bidirectional GRU \cite{cho-etal-2014-learning} with input embedding size 256 and hidden state size 512. The self-limited decoder is a two-layer GRU setting same input embedding size and hidden state size of 512. The random probability of data augmentation is set as 0.7 in pre-training and 0.5 in training. We choose the AdamW optimizer \cite{Loshchilov2017DecoupledWD} with weight decay $1e^{-2}$ and step decline schedule with decaying rate 0.5. During pre-training, the learning rate of language model is initialized as $5e^{-4}$ decaying at 1K, 2K and 3K epochs with a total 4k epochs. During training, all modules of PGPSNet train together with initial learning rate as $5e^{-5}$ for language model and $1e^{-3}$ for other modules, decaying at 160, 280, 360, 440 and 500 uniformly with a total 540 epochs. In addition, the batch size and dropout rate are set as 128 and 0.2 in all processes. \\
\textbf{Evaluation metrics.} \
We evaluate performance of geometric solvers at two levels: numerical answer and solution program. At each level, there are three evaluation patterns: \textit{Completion}, \textit{Choice} and \textit{Top-3}. In Completion, the symbolic solver gives the result searched for and the neural solver selects the first executable solution program. The Choice is defined as choosing the correct option from four candidates but selecting one randomly if answer is not in. The Top-3 computes the accuracy ratio of correct answer existing among top three solution candidates, less strict than Completion pattern. We set beam size to 10 being the same as \cite{Chen2021} for all evaluation manners. Noting that the performance of solution program is often lower than the actual result due to the equivalence of operation orders and diversity of solution strategies.  \\
\textbf{Datasets.} \
We conduct experiments on PGPS9K dataset split in two ways as introduced in Section 3.1, signified as Geometry3K and PGPS9K in the following experiments. To simplify experiments, we adopt the textual clauses generated from ground truth of diagram annotation for both symbolic solvers and neural solvers. The language model is pre-trained from scratch on our PGPS9K on account of huge gap from natural corpus and short of geometric corpus. Datasets GeoQA  \cite{Chen2021} and GeoQA+ \cite{cao2022} are not considered in experiments because they have no fine-grained diagram annotation and their diagrams are low-quality, hard to be parsed by existing diagram parsers.

\begin{table*}
\centering
\footnotesize
\renewcommand\arraystretch{1.2}
    \begin{tabular}{cccccccccc}
        \toprule
        \multirow{3}{*}{\makecell[c]{Self-limited \\ Decoder}} & \multirow{3}{*}{Data Aug} & \multirow{3}{*}{\makecell[c]{Structural \\ Clauses}} & \multirow{3}{*}{\makecell[c]{Pre-trained \\ LM}} & \multicolumn{3}{c}{Ans acc}  & \multicolumn{3}{c}{Prog acc} \\ 
        \cmidrule(lr){5-7} \cmidrule(lr){8-10}
          &  &  &  & Completion & Choice & Top-3 & Completion & Choice & Top-3  \\
         \midrule
        \ding{52} & \ding{56} & \ding{52} & \ding{56} & 32.5 & 52.2 & 57.6 & 27.2 & 47.3 & 53.1 \\
        \ding{56} & \ding{52} & \ding{52} & \ding{56} & 28.2 & 48.3 & 50.7 & 25.4 & 42.7 & 45.6 \\
        \ding{52} & \ding{52} & \ding{56} & \ding{56} & 36.6 & 59.5 & 62.4 & 33.9 & 52.8 & 58.6 \\
        \ding{52} & \ding{52} & \ding{52} & \ding{56} & 38.4 & 61.7 & 64.8 & 34.8 & 54.2 & 59.2 \\
        \ding{52} & \ding{52} & \ding{56} & \ding{52} & 48.1 & 67.5 & 71.4 & 45.4 & 62.0 & 68.1 \\
        \ding{52} & \ding{52} & \ding{52} & \ding{52} & \textbf{65.0} & \textbf{77.9} & \textbf{80.7} & \textbf{62.8} & \textbf{72.4} & \textbf{78.2} \\
        \bottomrule
    \end{tabular}
\caption{Ablation studies on Geometry3K.}
\label{ablation_study}
\vspace{-0.3cm}
\end{table*}

\subsection{Comparison with State-of-the-art Methods}
To evaluate our PGPSNet solver, we compare its performance with state-of-the-art GPS solvers InterGPS \cite{Lu2021}, NGS \cite{Chen2021} and Geoformer \cite{Chen2022}.\\
\textbf{Symbolic solvers.} The InterGPS is a rule-based symbolic solver which parses the problem text and diagram into formal language. Table \ref{performance_compare} displays the performances of Inter-GPS in three input modes with the best search strategy, where Inter-GPS(Predict) indicates that diagram and text formal language are predicted by its diagram and text parser, Inter-GPS(Diagram GT) indicates that text formal language is generated by its text parser but diagram formal language uses the ground truth, Inter-GPS (All GT) indicates that all utilize the ground truth. On Geometry3K, the results show that our PGPSNet outperforms Inter-GPS(Predict), achieves comparable performance as Inter-GPS(Diagram GT), and is inferior to InterGPS(All GT) in Completion. But in Choice, PGPSNet has surpassed all input modes and even gains a 2.0\% improvement over InterGPS(All GT). On PGPS9K, we implement the InterGPS(Diagram GT) by converting diagram annotation into diagram formal language and using its text parser. The PGPSNet shows more performance improvements in Completion and Choice, and result in Top-3 implies the improvement potential of our PGPSNet.\\
\textbf{Neural solvers.} NGS and Geoformer are two neural solvers and we re-implement them on our dataset. For full and fair comparison, we keep their pre-trained diagram encoder fixed, take both semantic clauses and textual problem as their text input, and employ the same data augmentation. Thanks to our good modal representation and effective modal fusion, our PGPSNet demonstrate superior performance improvements compared to baseline neural solver, NGS and Geoformer as displayed in Table \ref{performance_compare}. Besides, we note that there remains a huge performance gap between our solver and human expert, having much room for improvement.


\subsection{Ablation Study}
To demonstrate effects of different modules of PGPSNet, we conduct ablation studies on Geometry3K, taking the self-limited decoder, data augmentation, structural clauses and pre-training as ablation objects as displayed in Table \ref{ablation_study}.  The comparison between row 1 and row 4 shows that the data augmentation promotes GPS through adding geometric representation knowledge into the augmented data. By comparing row 2 to row 4, we find that the self-limited decoder improves performance of geometric reasoning, who limits feature and search space to reduce difficulty of model learning.  The language model, with structural and semantic pre-training, brings an amazing performance gain, especially in Completion with a 26.6\% answer accuracy improvement, as shown in row 4 and row 6. We also discover that structural clauses show less affect without pre-training compared row 3 with row 4, but obtain a remarkable improvement with pre-training as displayed in row 5 and row 6, revealing that basic connection relations can promote model's structure cognition by a befitting modal fusion approach. The performance trends of solution program are consistent with the numerical answer in all experiments.     

\begin{figure}[t]
    \begin{center}
    \includegraphics[width=0.95\columnwidth]{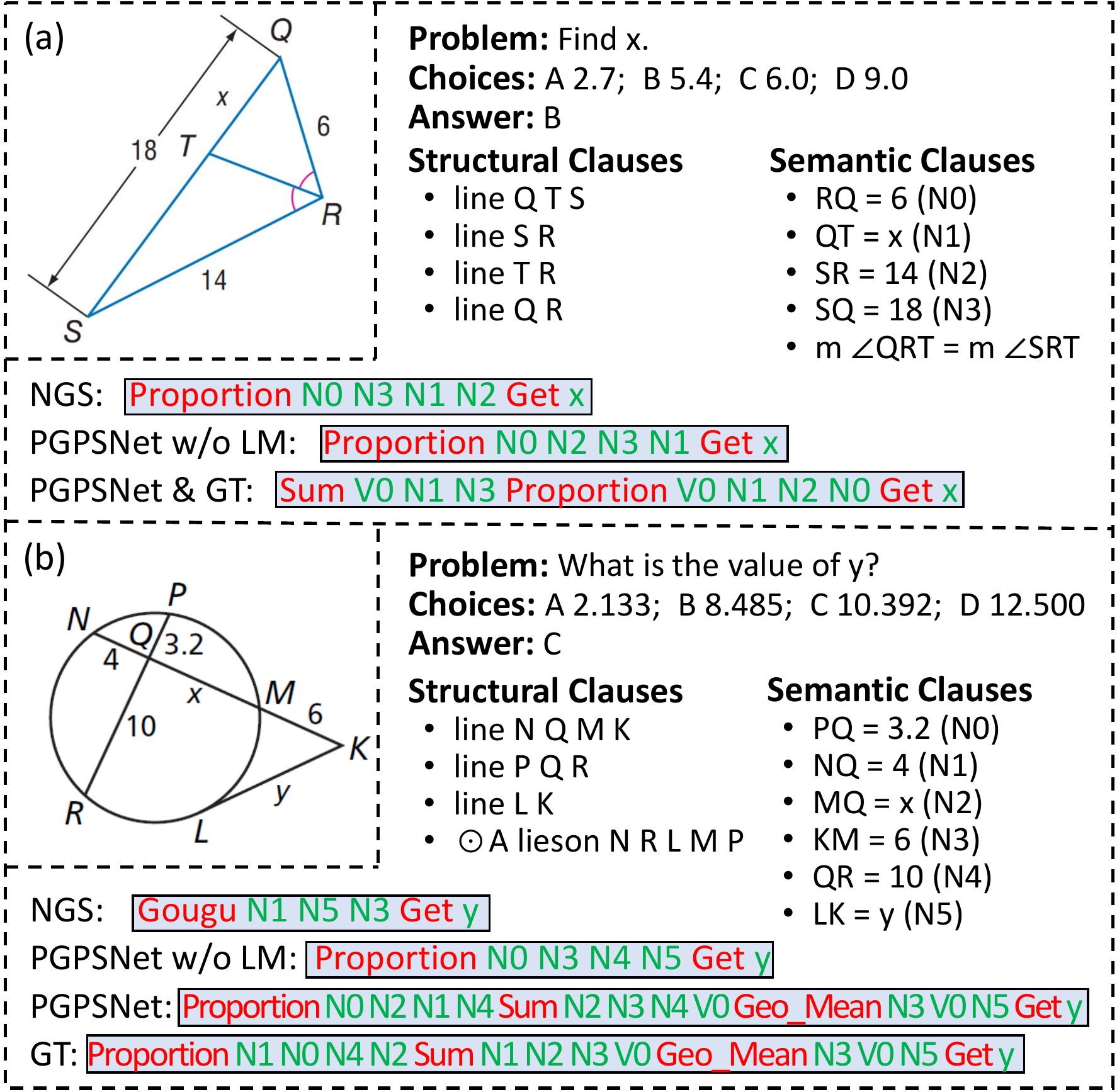} 
    \end{center}
    \vspace{-0.2cm}
    \caption{Case studies on PGPS9K.}
    \label{case_study}
    \vspace{-0.2cm}
\end{figure}

\subsection{Case Study}
 We also conduct case studies for discussing the ability and limitation of our PGPSNet as shown in Figure \ref{case_study}. The case (a) examines the application of Angle Bisector Theorem. The methods of NGS and PGPSNet w/o LM fail to determine the relation of corresponding edges correctly while our PGPSNet obtains the right solution program. The case (b) is challenging and need to combine two conditions of Segment Length Theorem. Solutions of all solvers are incorrect for case (b) but the PGPSNet's is most similar to the ground truth with only wrong in the second step, illustrating that PGPSNet is not yet qualified for complex geometric reasoning nowadays but has great potential.

\section{Conclusion}
In this work, we propose a new diagram-text fusion solver PGPSNet combining textual clauses parsed from diagram, and construct a large-scale and fine-annotated GPS dataset PGPS9K. Benefiting from effective modal representation and efficient modal fusion, PGPSNet makes full use of basic structural and semantic information to implement geometric reasoning. Besides, the interpretable solution program and well-designed data augmentations provide model with critical geometric knowledge for GPS such as geometry theorem and geometric representation. The experimental results demonstrate the potential of neural solvers and our work still has large room for improvement with more training samples or more elaborate modal fusion. 

\section*{Acknowledgments}
This work has been supported by the National Key Research and Development Program under Grant No. 2020AAA0109700, the National Natural Science Foundation of China (NSFC) grants U20A20223 and 61721004.

\bibliographystyle{named}
\bibliography{ijcai23}

\appendix


\section*{Appendix}

\section{Details about PGPS9K Dataset}
\subsection{Problem Types} 
\begin{figure*}
    \begin{center}
    \includegraphics[width=1.8\columnwidth]{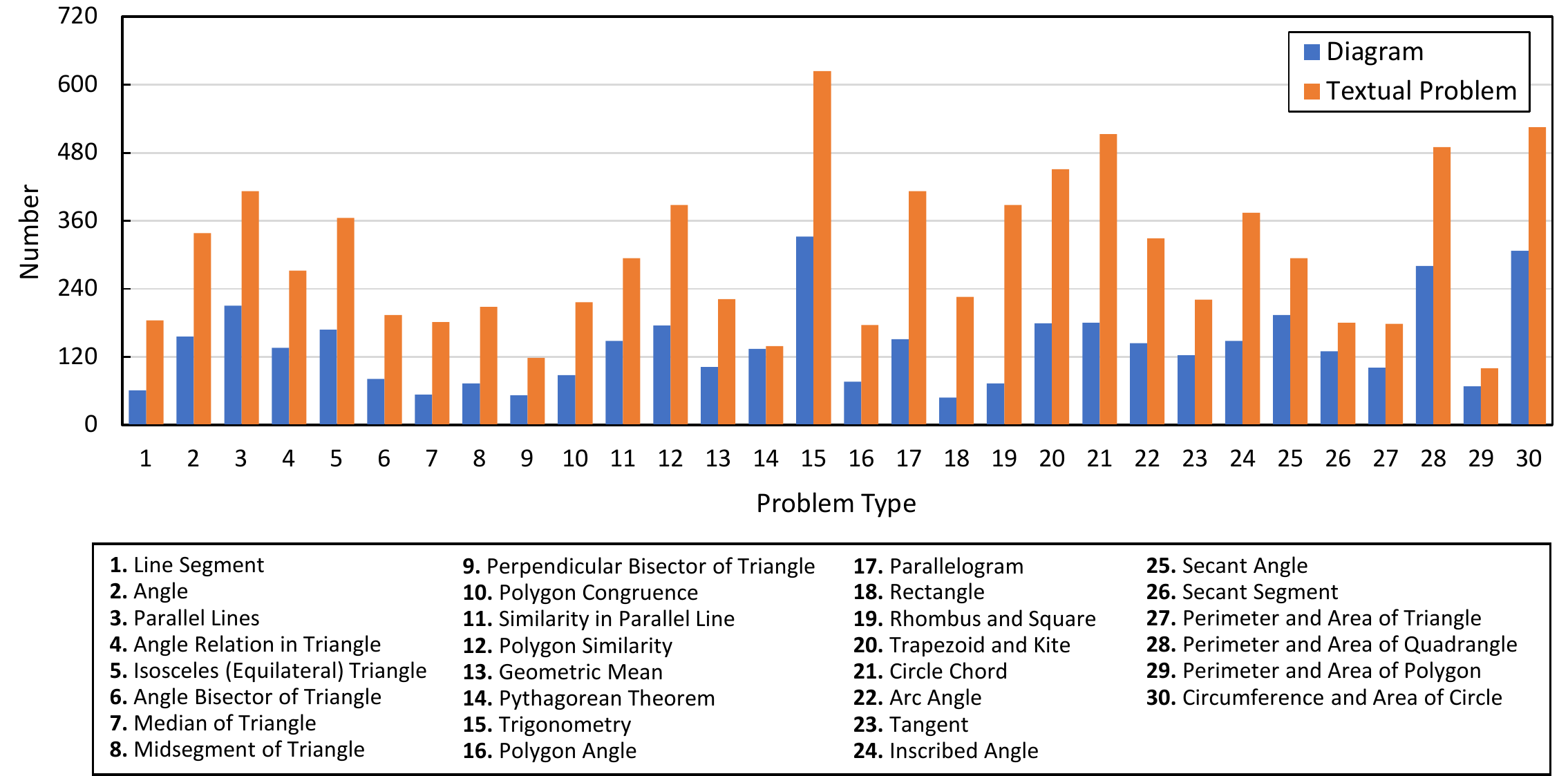} 
    \end{center}
    \caption{Distribution of problem types of PGPS9K dataset.}
    \label{problem_type}
\end{figure*}

Our PGPS9K dataset is divided into 30 problem types elaborately according to geometry knowledge points by education experts, covering almost all problem types of plane geometry problem across grade 6-12 as shown in Figure \ref{problem_type}. Fine-grained partition helps geometric problem collection, trying to keep relative balance of problem types. For example, the poor performance of some problem types may result from insufficient problem samples, so we could collect them on purpose. Moreover, fine-grained partition is beneficial to analyse the model performance of different problem types in depth. For instance, there widely exists intersection and inclusion of geometric patterns among different problem types, and exploring the model capacity of inductive and deductive on GPS is meaningful and interesting.

\subsection{Textual Clauses}
The textual clauses are linguistic descriptions of primitive relations in geometry diagram. They include structural clauses and semantic clauses, where the structural clauses present the connection relations of geometric primitives (point, line and circle) and the semantic clauses depict the relations between geometric primitives and non-geometric primitives (text and symbol). Table \ref{clause_template} displays the complete templates of textual clauses, consisting 3 types of structural clauses and 6 types of semantic clauses. These textual clauses are elementary and necessary which could be translated from relation tuples of diagram annotation directly without advanced geometry rules. The design of clauses is open but neural solvers do not pursue high-level logical clauses though they may contribute to problem solving.

\subsection{Solution Program}
Our annotation of solution program possesses better flexibility and scalability with extensive theorem operators and variable operands. The detailed contents of program set are listed in Table \ref{program_set}. It should indicate that solution program of geometry still confront similar issues as general math word problem: (1) \textit{Uncertainty caused by exchangeable operands.} The operands in some theorem formulas are commutative, e.g., in Pythagorean theorem with formula $a^2+b^2=c^2$, the two right edges are exchangeable. In our annotation, we normalize solution program via specifying two-level priority of commutative operands. The first is the class level that  "augment $>$  process variable $>$ problem variable $>$ constant" and the second is the index level with positive order. (2) \textit{Uncertainty of equivalent step orders.} Calculation steps are in no particular order sometimes. We keep the same pre-defined step order for the same problem type manually. (3) \textit{Multiple solution methods.} A part of geometry problems could be solved by multiple solution methods. We choose the solution method with the most concise solution program.  

\begin{table*}
\centering
\footnotesize
    \begin{tabular}{l|l|l|l} 
        \toprule
        \textbf{Type}  & \textbf{Template} & \textbf{Example} & \textbf{Explanation} \\ 
        \midrule
        \multirow{4}{*}{Structural Clauses} & line \& \& \& $\cdots$ & line A B C & \multirow{2}{*}{Points lie on line instances}  \\ 
        \cmidrule{2-3}
        & line * lieson \& \& \& $\cdots$ & line k lieson A B C & \\ 
        \cmidrule{2-4}
        & $\odot$\& lieson \& \& \& $\cdots$ & $\odot$O lieson E F G & Points lie on circle instances \\
        \midrule
        \multirow{8}{*}{Semantic Clauses}   &  \&\& = \&\& = $\cdots$ = \$ & AB = CD = $3x+y$ & Length of line segments \\ 
        \cmidrule{2-4}
        & l $\widehat{\&\&}$  = l $\widehat{\&\&\&}$ = $\cdots$ = \$ & l $\widehat{\rm{EF}}$ = $5\pi$ & Length of arcs \\ 
        \cmidrule{2-4}
        & m $\angle$\&\&\& = m $\angle$\& = m $\angle$\% = $\cdots$ = \$ & m $\angle$A = m $\angle 1$ = $30$ & Degree of angles \\ 
        \cmidrule{2-4}
        & m $\widehat{\&\&}$  = m $\widehat{\&\&\&}$ = $\cdots$ = \$  & m $\widehat{\rm{EFG}}$ = $270$ & Degree of arcs \\ 
        \cmidrule{2-4}
        & \&\&(line *) $\parallel$ \&\&(line *) $\parallel$ $\cdots$ ~  & line k $\parallel$ line m $\parallel$ EF & Parallel relation among lines \\ 
        \cmidrule{2-4}
        & \&\&(line *) $\perp$ \&\&(line *) on \& & EF $\perp$ GH on C & Perpendicular relation among lines  \\
        \bottomrule
    \end{tabular}
\caption{Templates of textual clauses. The symbols of '\&', '*', '\$', '\%' denote point, line, variable and angle ID, respectively.}
\label{clause_template}
\end{table*}

\begin{table*}
\centering
\footnotesize
\renewcommand\arraystretch{1.2}
    \begin{tabular}{p{1.4cm}l|l} 
    \toprule
    \multicolumn{2}{l|}{\textbf{Type}} & \textbf{Program Set} \\ 
    \midrule
    \multicolumn{2}{l|}{Operator} & \makecell[l]{Get, Equal, Sum, Multiple, Ratio,
      Median, Gougu, Gsin, Gcos, Gtan, \\ 
      Sin\_Law, Cos\_Law, Iso\_Tri\_Ang, Proportion, Geo\_Mean, Chord2\_Ang, \\
      TanSec\_Ang, Tria\_BH\_Area, Tria\_SAS\_Area, PRK\_Perim, Para\_Area, \\
      Rect\_Area, Rhom\_Area, Kite\_Area,  Trap\_Area, Circle\_R\_Circum, \\
      Circle\_D\_Circum, Circle\_R\_Area, Circle\_D\_Area, ArcSeg\_Area, \\
      Ngon\_Angsum, RNgon\_B\_Area, RNgon\_L\_Area, RNgon\_H\_Area}  \\ 
    \midrule
    \multirow{5}{*}{Operand} & Problem Variable & N0, N1, N2, $\cdots$, N10\\
              \cmidrule(lr){2-2}  \cmidrule{3-3}          
              & Process Variable & V0, V1, V2, $\cdots$, V6 \\
              \cmidrule(lr){2-2}  \cmidrule{3-3}  
              & Augment & a, b, c, $\cdots$, x, y, z \\
              \cmidrule(lr){2-2}  \cmidrule{3-3}  
              & Constant & C0.5, C2, C3, C4, C5, C6, C8, C60, C90, C180, C360    \\
    \bottomrule
    \end{tabular}
\caption{Program sets defined in our solution program, consisting of 34 operators and 55 operands, where operands involve 11 problem variables, 7 process variables, 26 augments and 11 constants.}
\label{program_set}
\end{table*}


\end{document}